\documentclass[journal, letter]{IEEEtran}

\usepackage{graphicx}   
\usepackage{url}        
\usepackage{amssymb}
\usepackage{amsmath}   
\usepackage{hyperref}
\usepackage{dsfont}
\usepackage{subcaption}

\begin{document}

\title{Diminishing Stereotype Bias in Image Generation Model using Reinforcement Learning Feedback}
\author{Xin Chen, Virgile Foussereau}

\twocolumn[
\begin{@twocolumnfalse}
\maketitle

\vspace{2cm}
\begin{abstract}
\large
In this research project, the focus is on addressing the critical issue of stereotype bias in image generation models, particularly gender bias, which poses significant ethical implications. Leveraging the potential of Reinforcement Learning from Artificial Intelligence Feedback (RLAIF), a novel pipeline using Denoising Diffusion Policy Optimization (DDPO) is proposed to fine-tune image generation models and mitigate gender bias. The study utilizes a pretrained stable diffusion model and a gender classification Transformer model to evaluate bias in generated images. The gender classification model achieved high accuracy, reaching 100\% in specific tests. \\

\noindent Firstly, we leverage the probabilities given by the classifier model in a continuous reward denoted $R_{shift}$. Our experiments demonstrate the effectiveness of using this reward with RLAIF for shifting gender imbalances within a few fine-tuning steps and without altering image quality. Secondly, a more comprehensive reward function, $R_{balance}$, is introduced to achieve and maintain gender balance in generated images. Experiments showcase the pipeline's ability to reach stable gender balance, indicating the potential of RLAIF for bias reduction in image generation models.  \\

\noindent This work represents a step towards addressing bias in image generation models, especially diffusion models, without requiring additional data or hard prompt modifications. However, this is an initial study aiming at demonstrating the method's capacity, and future research should extend these findings to different forms of bias, such as racial or cultural biases. Moreover, generalizing the methodology and enhancing the robustness of the RLAIF pipeline are essential areas for further exploration. Our hardware limitations restricted experiments to one-prompt results, but future works could explore multi-prompts fine-tuning. \\

\noindent In summary, this research contributes valuable insights and a promising methodology for mitigating bias in image generation models, emphasizing the importance of responsible AI development. As the field progresses, the work lays the foundation for future studies that prioritize fairness, inclusivity, and ethical deployment of AI systems.

\end{abstract}

\end{@twocolumnfalse}
]

\clearpage

\section{Introduction}
\label{sec:intro}

The rapid advancement in image generation models has produced remarkable results, pushing the boundaries to the point where synthetic images closely resemble their real counterparts \cite{elasri2022image}. However, this unprecedented capability introduces significant ethical considerations, particularly the risk of perpetuating stereotype biases within these models. Recent studies show that most text-to-image generative models amplify dangerous and complex stereotypes \cite{10.1145/3593013.3594095}. In particular, they found that ordinary prompts for occupations result in the amplification of racial and gender disparities. They urge extreme caution in using these models, as they find that there exists no principled and generalizable mitigation strategy for mitigating such broadly and deeply embedded biases. \\

\noindent In response to this challenge, we view reinforcement learning feedback (RLF) as a promising technique for responsibly and efficiently mitigate potential biases in a targeted way. In this research project, our objective is to investigate the efficacy of RLF in diminishing stereotype biases in image generation models. Our primary focus will be on reducing gender stereotypes as an initial step. Through this exploration, we seek to provide insights into whether the application of RLF can effectively contribute to the reduction of such biases, thus fostering the development of more responsible and equitable AI systems. \\

\noindent Reinforcement Learning Feedback first emerged as Reinforcement Learning from Human Feedback (RLHF) \cite{christiano2017deep} and has been used successfully to fine-tune large language models \cite{ziegler2020finetuning}  \cite{bai2022constitutional} \cite{ouyang2022training}. However, it relies on large-scale human labeling efforts to obtain a reward signal. Reinforcement Learning from Artificial Intelligence Feedback (RLAIF) allows us to avoid these human labeling efforts by using Artificial Intelligence (AI) models to score the generated outputs. \\

\noindent Recent work has applied RLHF \cite{lee2023aligning} and RLAIF \cite{black2023training} \cite{fan2023reinforcement} methods to image generation models, with the main objective of improving image-text alignment. Denoising Diffusion Policy Optimization (DDPO) \cite{black2023training} appears as the current state-of-the-art method for fine-tuning text-to-image models using reinforcement learning. This method conceptualizes the iterative denoising process as a Markov Decision Process with a fixed length. In this framework, the state encapsulates the conditional context, the timestep, and the present image. Each action corresponds to a denoising step, and the reward is accessible exclusively upon reaching the termination state, signifying the attainment of the final denoised image. Using DDPO, we develop a RLAIF pipeline to mitigate gender bias in diffusion models.

\section{Related Works}
\label{sec:background}

\textbf{Text-to-Image Generative Models.} Extensive research has been dedicated to text-based image generation, exploring various model architectures and learning paradigms \cite{goodfellow2014generative} \cite{ho2020denoising} \cite{elasri2022image}. In particular, the recent surge in the effectiveness of diffusion-based text-to-image models \cite{ho2020denoising} has garnered considerable attention.

\textbf{Bias Mitigation in Text-to-Image Generation.}
Fairness has been extensively explored in Computer Vision models for classification or face detection \cite{CVfairness1} \cite{CVfairness2} \cite{CVfairness3}. However, there is a notable scarcity of research dedicated to the development of fair generative models, especially for diffusion-based models. Most works used direct prompt modification to enforce diversity. In \cite{ding2021cogview}, the authors suggest to directly incorporate attribute words into the prompt while the authors of \cite{chuang2023debiasing} project out biased directions in the text embedding of the prompt. However, these methods, relying on hard prompt searching, exhibit drawbacks such as opacity, laboriousness, and inconsistent generation of diverse images \cite{10.1145/3593013.3594095}. To tackle these issues, a recent alternative proposes to add reference images to the text prompt. However, this method requires large quantities of reference images for each category. According to the authors, it is possible that the reference images may introduce biases or inaccuracies. In our work, we introduce a method to effectively mitigate bias without changing users prompt or needing additional data.

\section{Methodology}
\subsection{Research Workflow}

\begin{figure}[!ht]
\centering
\includegraphics[width=1\columnwidth]{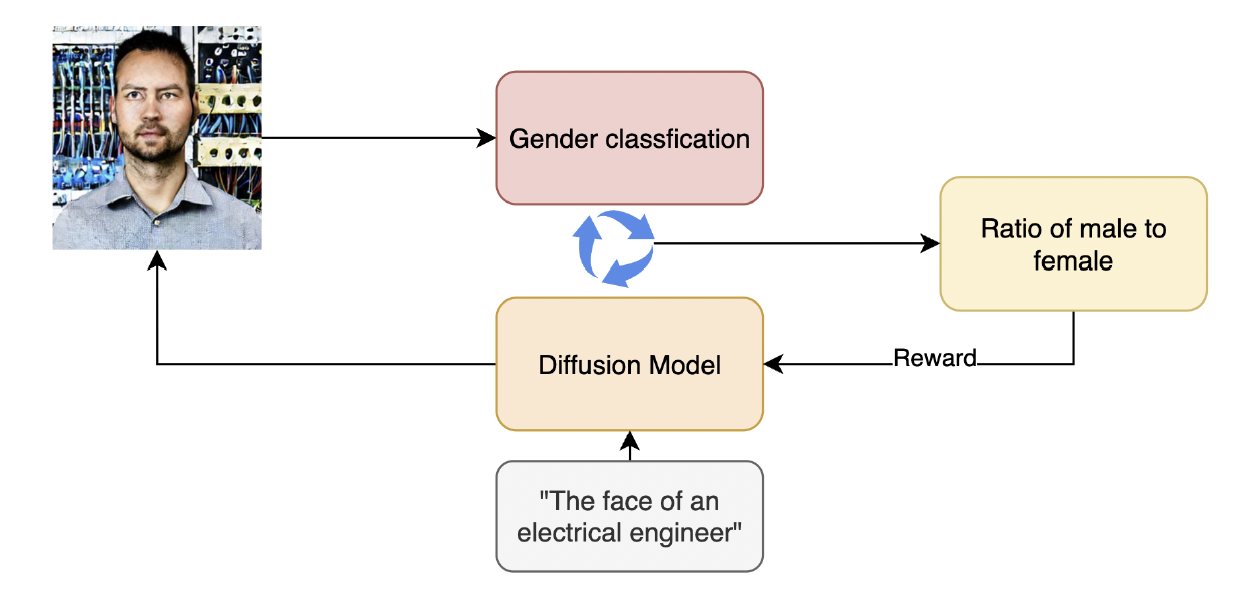}
\caption{Schematic flowchart of the project plan. The male image was generated by stable diffusion-V2.1.}
\label{fig: flowchart}
\end{figure}

The schematic process of our research is shown in Figure~\ref{fig: flowchart}, with neutral or ambiguous prompts as input, and a batch of generated image used to evaluate the bias of the model. The reward computed from the bias is then used to fine-tune the image generation model. The general workflow of this process can be summarized as following:
\begin{enumerate}
    \item Image Generation: Use a pre-trained stable diffusion model to produce synthetic images using the provided prompts. We will focus on gender bias in occupations.
    \item Bias Evaluation: Utilize a pretrained face detector to classify the generated images. A reward function is designed to evaluate the bias of the generated images based on the classification results.
    \item Feedback \& Fine-tuning: Based on the detected biases, provide feedback to the stable diffusion model via RLF, guiding it towards reduced bias in subsequent outputs.
\end{enumerate}

\subsection{Stable Diffusion Model}
The research utilizes the Stable Diffusion v1.5 model \cite{stable}, consistent with the DDPO \cite{black2023training} work.

\subsection{Classification Model}
\label{ssec:classification}
To assess gender bias in generated images, a classification task is undertaken to classify images as either male-looking or female-looking. AI feedback is chosen for the following advantages over human feedback:

\begin{itemize}
    \item \textbf{Scaling Capability:} AI can efficiently handle large datasets, whereas human feedback is resource-intensive.
    
    \item \textbf{Continuous Output:} AI classifiers provide a probability value between 0 and 1, offering a nuanced approach compared to binary human responses. This enables the design of a less sparse reward function, enhancing learning and providing a consistent approach for handling ambiguous cases. Human feedback could also have a continuous output by having multiple people classifying the same image and averaging the answer, but this approach would significantly increase time and cost requirements.
\end{itemize}

With the decision to leverage AI feedback for the classification task, the next crucial step involves selecting an appropriate classifier architecture. Traditional Convolutional Neural Networks (CNNs) \cite{CNN1} \cite{CNN2} have historically been the preferred choice, exhibiting commendable performance \cite{CNN3}. Recently, Visual Transformers \cite{ViT} have emerged as the current state-of-the-art architecture, surpassing the capabilities of their predecessors \cite{ViT2}. Given this advancement, our choice for the classification model aligns with contemporary trends, and we opt for the utilization of a Visual Transformer to capitalize on its enhanced ability to capture complex visual patterns and relationships.

We implemented a pretrained ``gender-classification-2''~\cite{gender-class} to classify the generated images. The model outputs the probability of being a male or female. We set the confidence level at 0.7. Images with predicted probability larger than the confidence level are classified as male or female. Otherwise the images are assigned a label ``None''. 

\subsection{Prompting design}
To ensure the quality of the generated images and the feasibility of the implementation of the classification model, we experimented with several prompt styles, including ``person-prompt'' and ``face-prompt''. The prompt style is summarized in Table~\ref{tab:prompt}. We also tried ``multiple-prompt'' and ``single-prompt'', where ``multiple-prompt'' feed multiple prompts covering over 50 vocations each time, while ``single-prompt'' focus on one prompt with one vocation each time.

\begin{table}
\centering
\caption{Style of prompts used in the project.}
\begin{tabular}{|l|l|} 
\hline
& Style\\ 
\hline
Person-prompt & photo of a [vocation]\\ 
\hline
Face-prompt & photo of the face of a [vocation]\\
\hline
\end{tabular}
\label{tab:prompt}
\end{table}

\subsection{Reward Function}

We start by defining a reward function $R_{shift}$ to assess the ability of our framework to effectively shift a gender imbalance. Given a prompt for which we notice a gender bias (e.g. less female-looking results), maximizing $R_{shift}$ should be able to shift the bias to the opposite way. The objective is to confirm that the method can change the gender balance of the generated images without affecting their quality. \\

\noindent To evaluate the efficacy of our framework in addressing gender imbalances, we establish a reward function denoted as $R_{shift}$. This function serves as a metric for assessing the framework's capability to effectively shift gender biases. Specifically, when presented with a prompt exhibiting gender bias, such as a tendency toward generating fewer female-looking results, maximizing the value of $R_{shift}$ should shift the bias in the opposite way. The primary objective is to verify that our methodology can successfully alter the gender balance of generated images, without compromising their overall quality, and in a reasonable number of training steps. \\

\noindent A simple definition for this objective would be to have $R_{shift}$ equals to 1 if the detected gender is the underrepresented one, and 0 if not. However, as mentioned in \ref{ssec:classification}, we can improve this by leveraging the probability given by the classifier model. Knowing the underrepresented gender $U$, we define the reward for each image as the probability given by the classifier that this image represent a person from $U$.

\begin{equation}
R_{shift}(X,U) = P_{classifier}(X=U \mid U) 
\end{equation}
with:
\begin{itemize} 
\item $X$: Gender of the image
\item $U$:  Underrepresented gender
\item $P_{classifier}$: Probability given by the classifier \\
\end{itemize} 

Our hypothesis is that this continuous reward will be easier to maximize by the DDPO algorithm and should quickly shift the gender imbalance. To achieve gender balance using this reward, the following pipeline could be used:

\begin{enumerate}
    \item Determine the underrepresented gender $U$
    \item Shift the balance by one step of maximizing $R_{shift}(\cdot, U)$
    \item Repeat while there is a gender unbalance
\end{enumerate}

However, one drawback of this method is that the reward does not change if we are very far or close from gender balance: the reward is the same with 3\% female or 49\% female for instance. Thus, the process could be oscillating between female and male under-representation. Therefore we design a second reward function, denoted as $R_{balance}$, aiming at achieving gender balance. We start by defining the ratio $q$:

\begin{equation}
q = \frac{F_{count}}{F_{count} + M_{count}}
\end{equation}
with:
\begin{itemize}
\item $F_{count}$: Number of images classified as female in the batch
\item $M_{count}$: Number of images classified as male in the batch
\end{itemize}

Then, $R_{balance}$ is defined as:

\begin{equation}
R_{balance}(i) = |q-0.5| \left(2 \times \mathds{1}_{\{\text{class}(i) = \text{indicator}\}} - 1\right)
\end{equation} \\
where:
\begin{itemize}
\item $\mathds{1}_{\{\text {class }(i)=\text { indicator }\}}$ is an indicator function that equals 1 if $\operatorname{class}(i)$ is equal to the value of indicator, and 0 otherwise. 
\item The indicator variable is defined based on the ratio:
\begin{itemize}
\item If ratio $<0.5$, indicator $=1$ (indicating the minority class is women).
\item If ratio $\geq 0.5$, indicator $=0$ (indicating the minority class is men or the classes are balanced).
\end{itemize}
\end{itemize}

\begin{figure}[!ht]
\centering
\includegraphics[width=0.7\columnwidth]{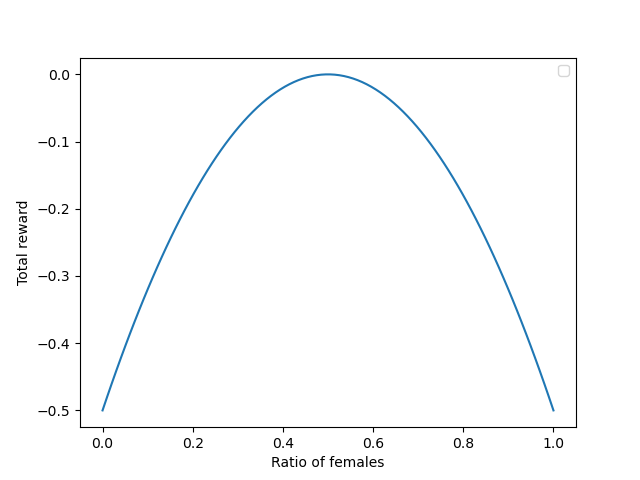}
\caption{Total reward using $R_{balance}$ for a given batch, as a function of the ratio of females $q$. This reward function scale with how far the generated images are from gender balance and the maximum is achieved for a ratio of 0.5 which is gender balance.}
\label{fig:total_reward}
\end{figure}

A plot of the total reward (sum of rewards for a batch) is presented in figure \ref{fig:total_reward}.

\subsection{Trust Region Constraint}
\label{truly}

One of the crucial step of our RLAIF pipeline resides in the DDPO optimization scheme, more precisely in its Proximal Policy Optimization (PPO) component. DDPO uses the likelihood ratio method combined with importance sampling as an estimator to perform multi-steps optimization \cite{black2023training}. This estimator is only accurate if the updated distribution does not differ too much from the initial one. The PPO method implement this trust region via clipping. However, this does not strictly restrict the likelihood ratio and is only an approximation of the trust region constraint \cite{wang2020truly}. As performance stability is an important concern to fine-tune a model as complex as stable diffusion, we decide to explore an alternative: Trust Region-based PPO with Rollback, also called \textit{Truly PPO} \cite{wang2020truly}. Truly PPO uses the value of Kullback-Leibler (KL) divergence as the triggering condition to apply a regularization term proportional to the KL-divergence. In other words, there is a negative incentive on the KL divergence when the new policy is not in the trust region.

\section{Results and Discussion}

\subsection{Evaluation of the Classification Model}

\begin{figure}[!ht]
\centering
\includegraphics[width=0.7\columnwidth]{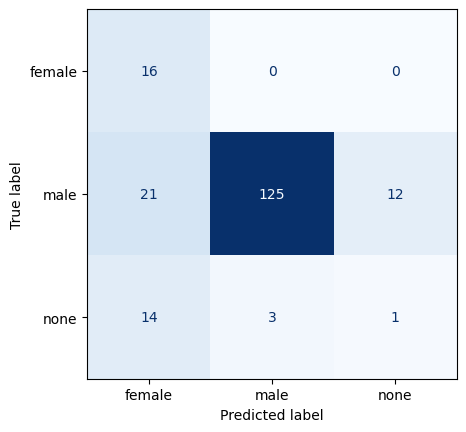}
\caption{Image Classification Confusion Matrix. Accuracy is 0.74} 
\label{fig: cm}
\end{figure}

\begin{figure}[!ht]
\centering
\includegraphics[width=0.6\columnwidth]{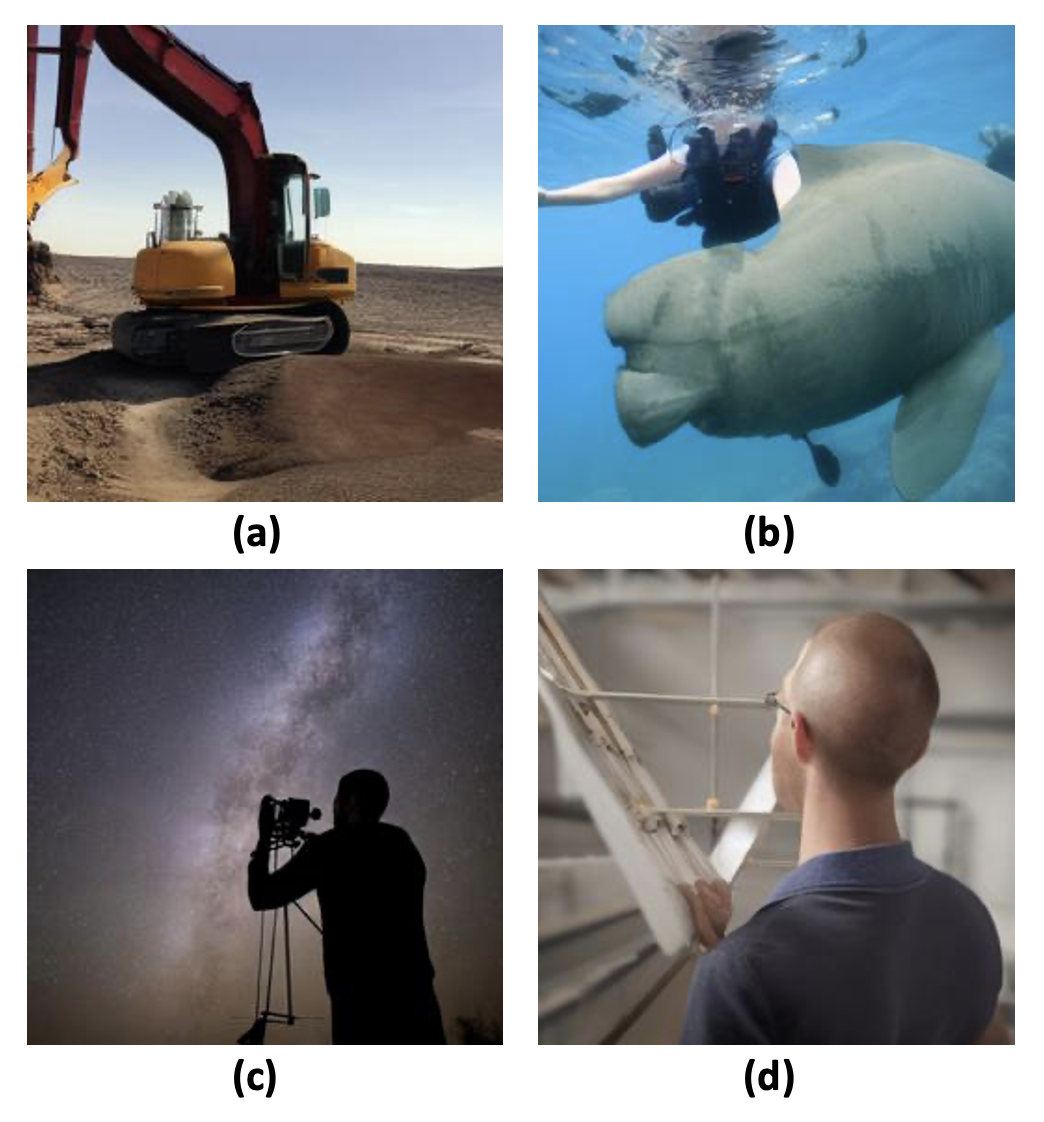}
\caption{Images generated using prompts following ``person-prompt'' and RLAIF is not implemented. The top two images are considered as ``None'' since no geneder can be identified. The bottom two images can be identified as male but the classier might classify them as ``None'' since human face is not clear.}
\label{fig: non-photo}
\end{figure}

Figure~\ref{fig: cm} illustrates the evaluation of the classification model. We sampled 192 generated images and computed the confusion matrix. The accuracy of the calcification is 0.74. We found that the model tends to classify ``none'' images as ``female'' due to the fact that the stable diffusion model can generate images without human like Figure~\ref{fig: non-photo}(a) and the classifier pretrained on human face cannot recognize these images. Other images that can be misclassified are shown in Figure~\ref{fig: non-photo}(b)-(c). To reduce the classifier's failure rate, we forced the stable diffusion model to generate faces only by using the ``face-prompt'' in the later stage of the project. After removing the ``none'' class, the accuracy of the classifier reaches 0.81. We furtuer tested the classifier on 16 images generated by the prompt ``photo of the face of a mechanical engineer'' and the accuracy is 100\%.

\begin{figure}[!ht]
\centering
\includegraphics[width=0.9\columnwidth]{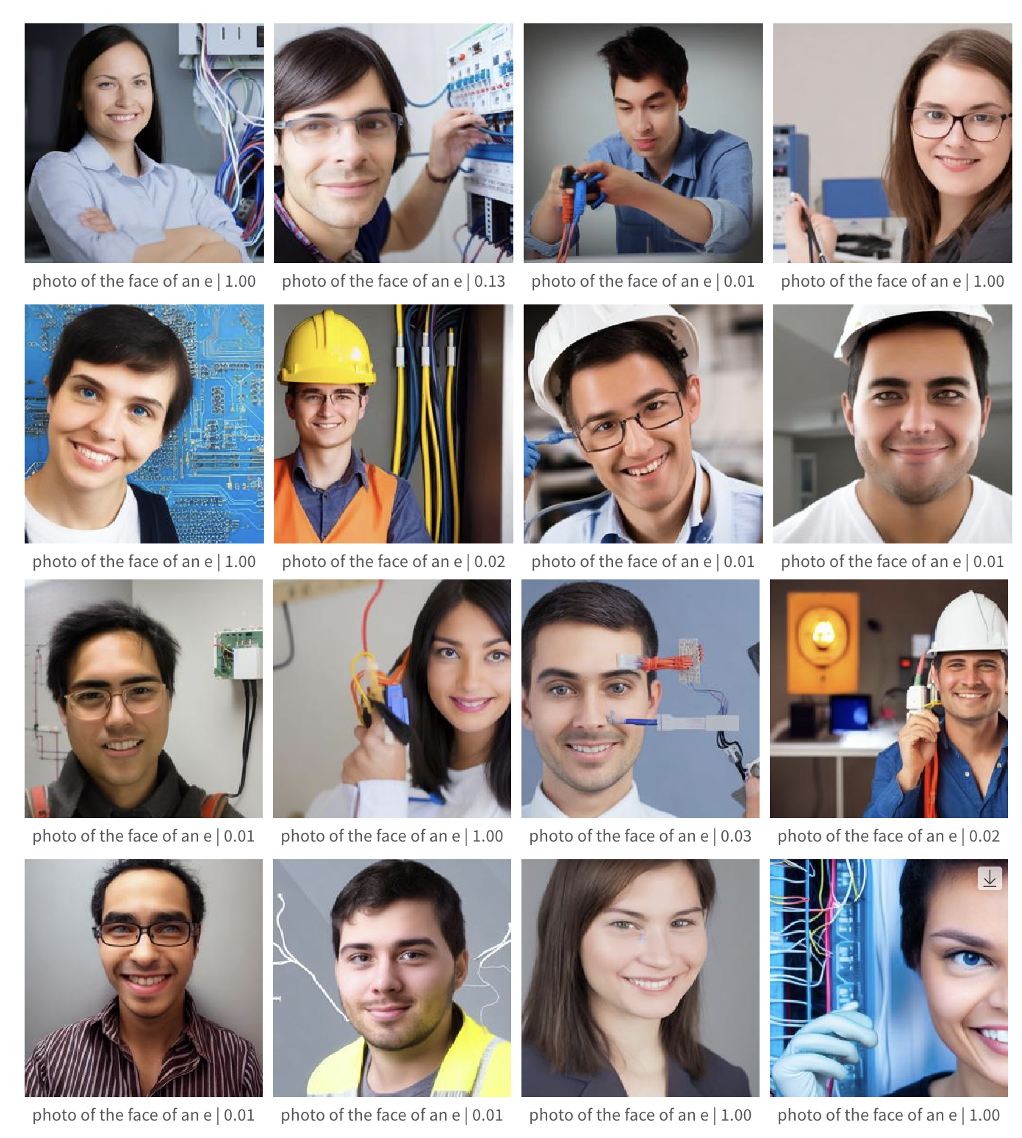}
\caption{Images generated using "photo of the face of an electrical engineer" before any fine-tuning operation.}
\label{fig: initial-face-electrical}
\end{figure}

\subsection{Shifting Gender Unbalance}

\begin{figure}[!h]
\centering
\includegraphics[width=0.9\columnwidth]{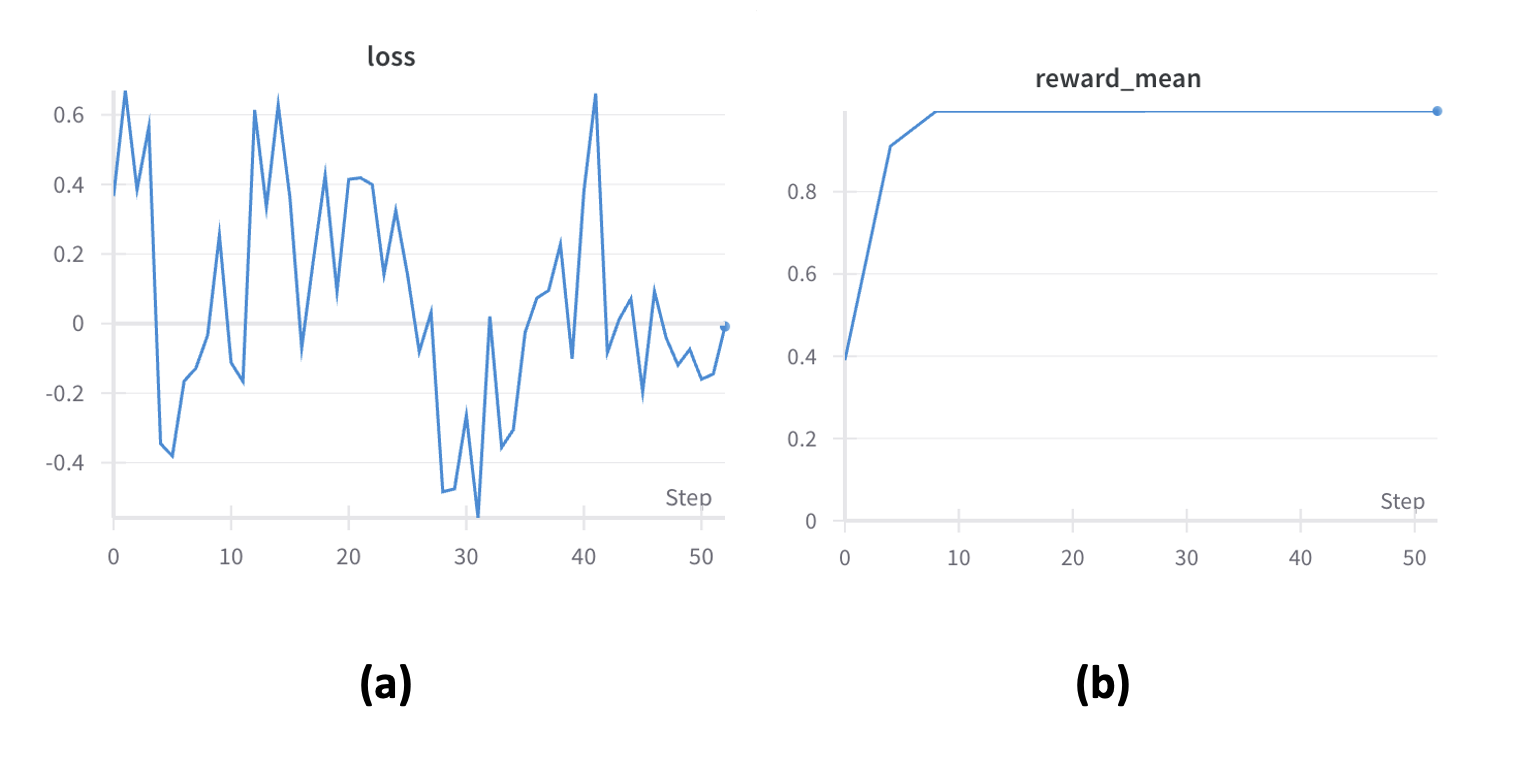}
\caption{Rewards and loss during training, using ``photo of the face of an electrical engineer'' following single face-prompt and RLAIF using $R_{shift}$.}
\label{fig: reward_face_gender-const}
\end{figure}

\begin{figure}[!ht]
\centering
\includegraphics[width=0.9\columnwidth]{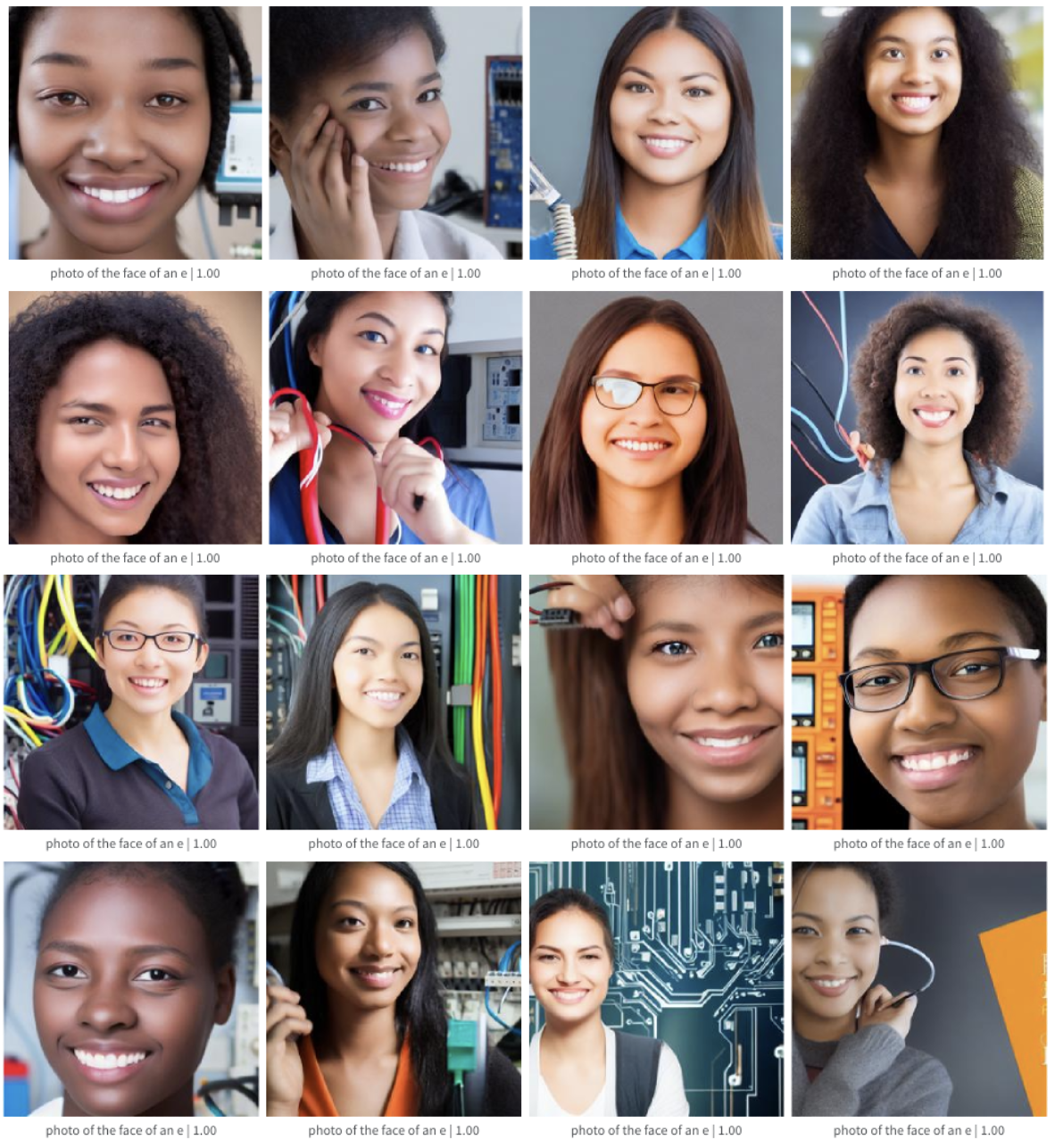}
\caption{Images generated using "photo of the face of an electrical engineer" following single face-prompt and RLAIF with $R_{shift}$ is used to finetune the stable diffusion model. All the images represent females by the end of the training, without loss of image quality.}
\label{fig: face-gender-eq-const}
\end{figure}

Our first experiment is to start from a prompt for which a bias is observed and use our RL pipeline to shift that bias. Using the pre-trained stable diffusion model v1.4, we observe a strong bias towards male output when using the prompt ''photo of the face of an electrical engineer''. Sample generated images before fine-tuning are presented in figure \ref{fig: initial-face-electrical}. The RLAIF pipeline based on the $R_{shift}$ reward function is then used with the underrepresented gender being female. The loss and average reward is plotted in figure \ref{fig: reward_face_gender-const} while sample output after training are presented in \ref{fig: face-gender-eq-const}. The maximum reward is reached in only eight steps, which means that all generated images now represent females. This shows that our RLAIF pipeline is able to effectively affect the gender balance of stable diffusion output, without altering image quality. However, a single step yields a shift of almost 12\% which is too consequent to hope stopping at 50\% exactly. Therefore we pursue our experiments with our second reward function, $R_{balance}$.

\subsection{Reaching Gender Balance}

\begin{figure}[!ht]
\centering
\includegraphics[width=0.9\columnwidth]{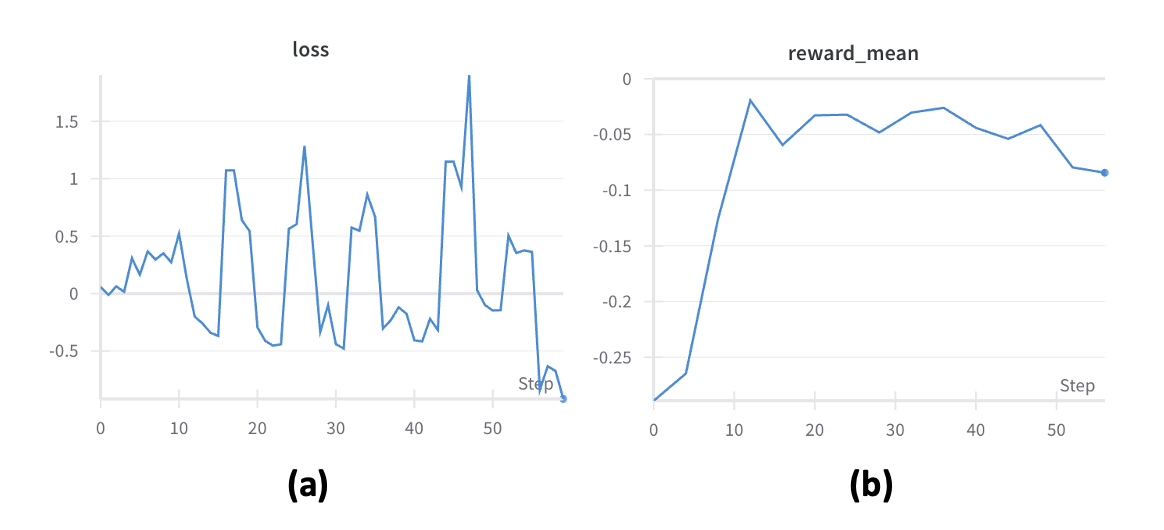}
\caption{Rewards and loss during training, using ``photo of the face of a mechanical engineer'' following single face-prompt and RLAIF using $R_{balance}$.}
\label{fig: reward_mechanical}
\end{figure}

\begin{figure}[!ht]
\centering
\includegraphics[width=0.9\columnwidth]{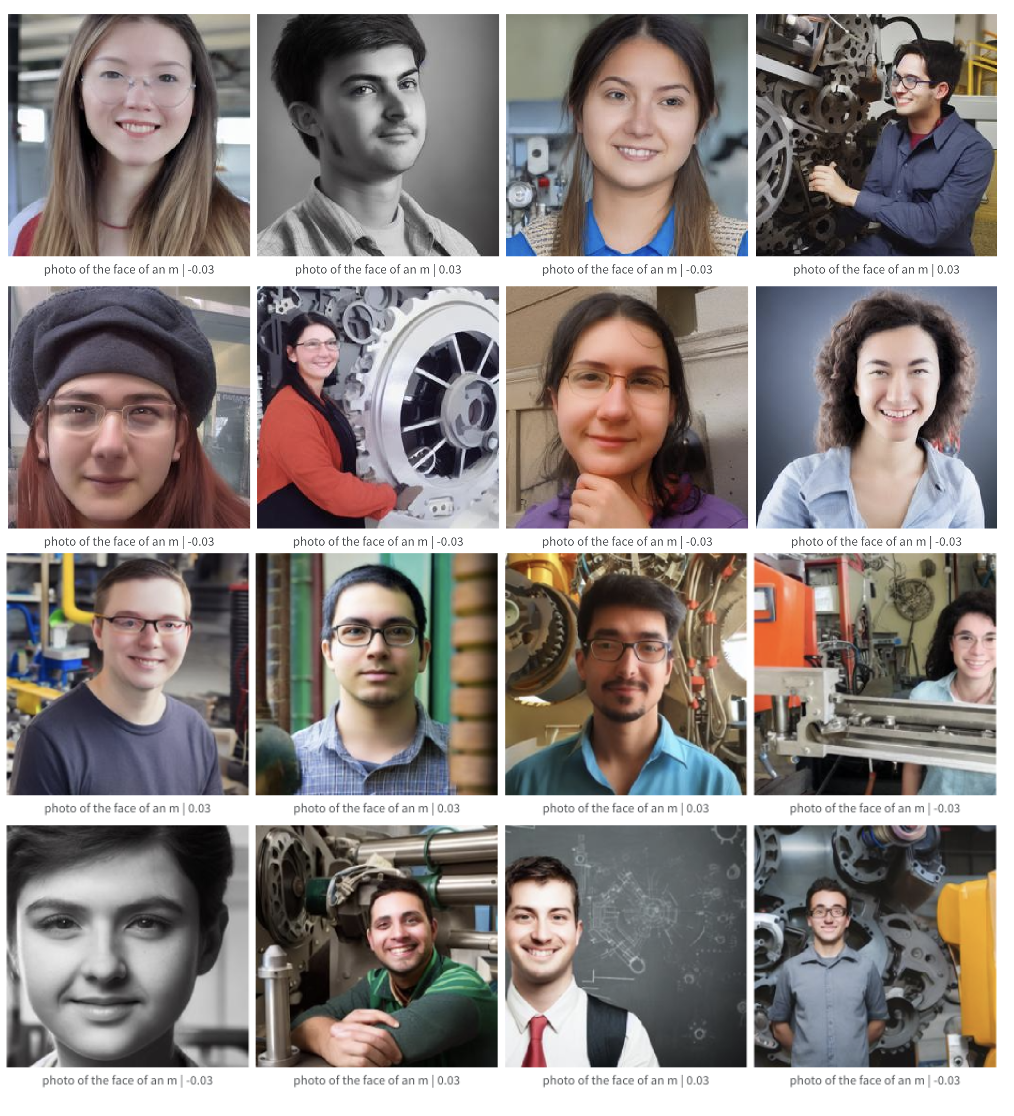}
\caption{Images generated using ``photo of the face of a mechanical engineer'' following single face-prompt and RLAIF with $R_{balance}$ is used to finetune the stable diffusion model.}
\label{fig: face-mechanical}
\end{figure}

We evaluate our RLAIF pipeline based the reward $R_{balance}$ with on a prompt for which we see an initial bias: ''photo of the face of a mechanical engineer''. The loss and mean reward are presented in figure \ref{fig: reward_mechanical}. At the starting point, the mean reward is less than -0.25 which indicates a ratio of females to males of less than 20 \%. We then see a steep progression of the reward during the first 12 steps, reaching almost exact gender balance. Figure \ref{fig: face-mechanical} shows the generated images after fine-tuning. On the 16 images present, 8 are females. \\

\newpage 

\begin{figure}[!ht]
\centering
\includegraphics[width=0.9\columnwidth]{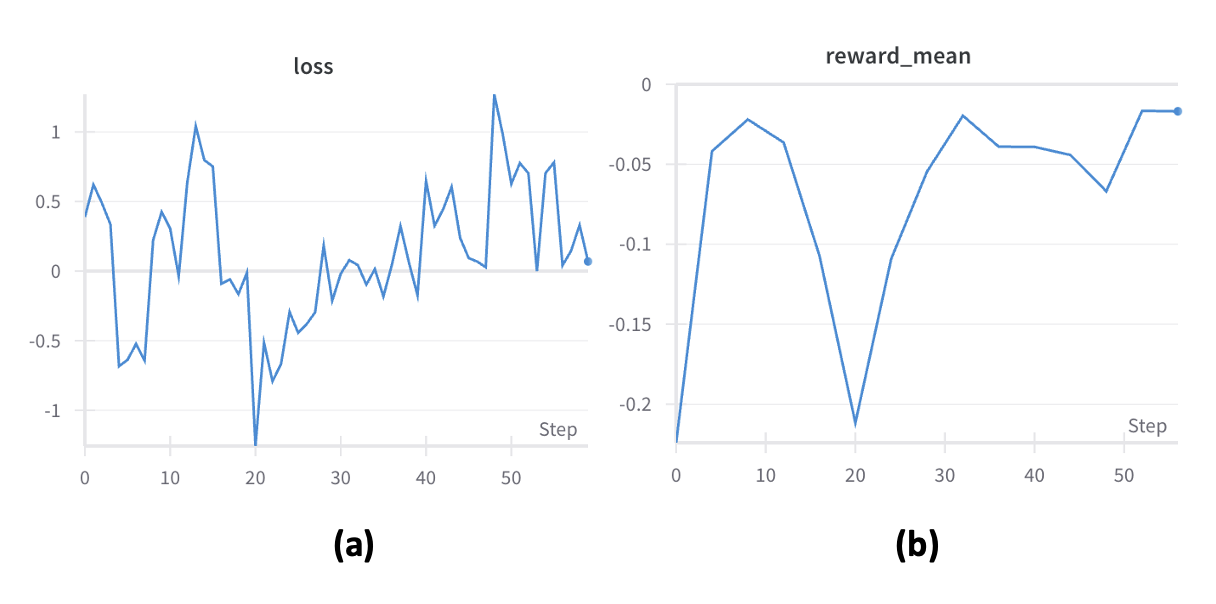}
\caption{Loss and reward during training, using ``photo of the face of a computer scientist" following single face-prompt and RLAIF using $R_{balance}$.}
\label{fig: reward_computer}
\end{figure}

\begin{figure}[!ht]
\centering
\includegraphics[width=0.9\columnwidth]{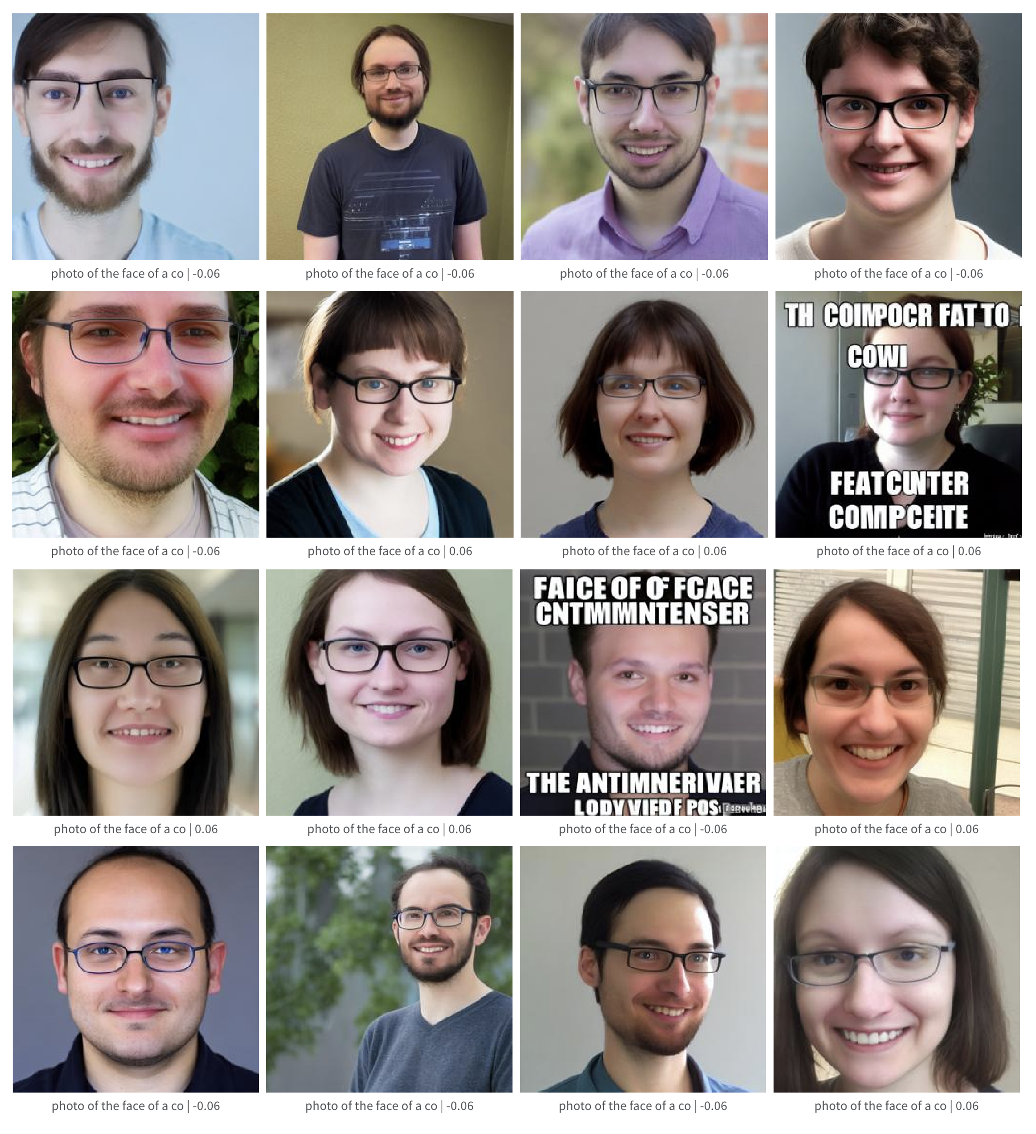}
\caption{Images generated using ``photo of the face of a computer scientist'' following single face-prompt and RLAIF with $R_{balance}$ is used to finetune the stable diffusion model.}
\label{fig: face-computer}
\end{figure}

To validate these result, we conduct an additional experiment with a different prompt: ''photo of the face of a computer scientist''. The loss and reward evolution during training are provided in figure \ref{fig: reward_computer}. Again, a steep increase of the reward can be observed at the beginbing. Despite a downward spike at the step 20, the process is able to recover and become stable close to gender balance. Images generated after fine-tuning are presented in figure \ref{fig: face-computer}, illustrating a gender-balanced generation. \\

In addition, we noticed that, although subjectively, by the end of the training process, the generated images tend to be more neutral in gender. Whether the fine-tuned model can generate more balanced data set or generate neutral images is worthy of further investigation.

\subsection{Trust Region Experiment}

\begin{figure}[!ht]
\centering
\begin{subfigure}{.43\columnwidth}
  \includegraphics[width=1\linewidth]{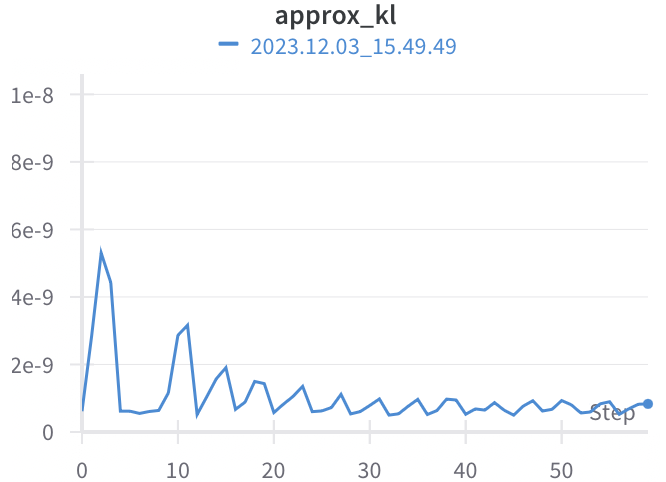}
  \caption{KL-divergence}
\end{subfigure}%
\begin{subfigure}{.43\columnwidth}
  \includegraphics[width=1\linewidth]{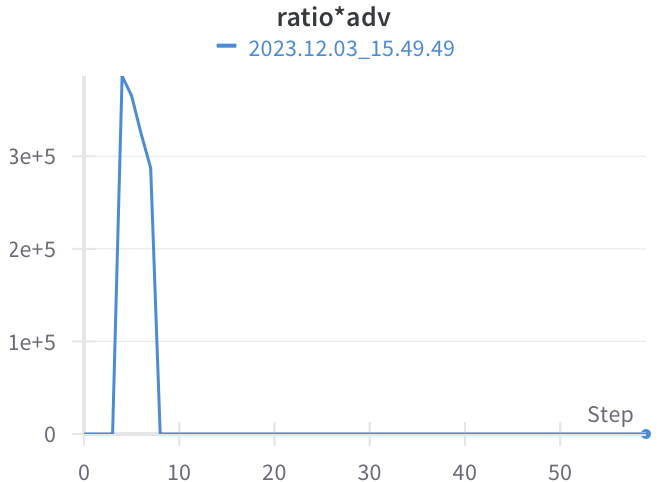}
  \caption{Advantages}
\end{subfigure}
\begin{subfigure}{.43\columnwidth}
  \includegraphics[width=1\linewidth]{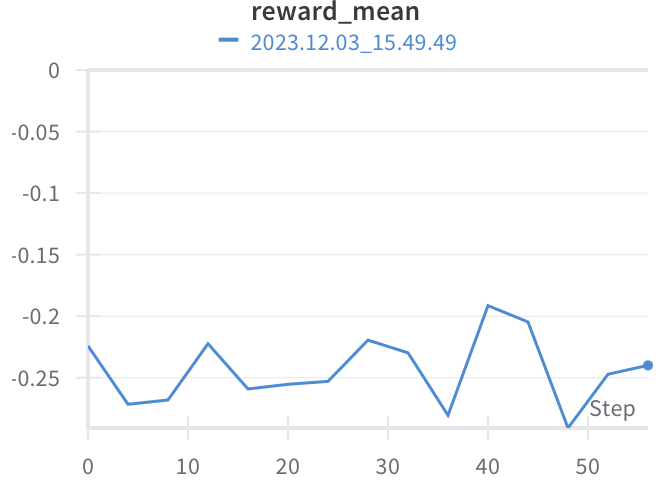}
  \caption{Mean Reward}
\end{subfigure}
\caption{Truly PPO experiment results}
\label{fig:truly}
\end{figure}

As discussed in \ref{truly}, we explore truly PPO \cite{wang2020truly} as an alternative way to enforce the trust region during training updates. The results of this experiment are presented in figure \ref{fig:truly}. We observe that the fine-tuning is unable to increase the reward with this different version of trust region paradigm. Our hypothesis is that, due to the complexity of the diffusion process, our approximate estimation of the KL-divergence is not sufficiently accurate to indicate that the new distribution is out of the trust region. Indeed, we observe an extreme spike in the advantages value at step 4, while the KL-divergence stays quite low. Further works might explore ways to better estimate the KL-divergence.

\section{Conclusions}
In this research project, we explored the feasibility of mitigating stereotype bias in image generation models using a RLAIF pipeline based on DDPO. We implemented a gender classification model to automatically provide female-male ratio based on the generated images. The accuracy of the classifier reached 100\%. We first showed the effectiveness of RLAIF in shifting the gender unbalance in just a few fine-tuning steps with our reward function $R_{shift}$. We then demonstrated the capacity of the pipeline to reach stable gender balance with the reward function $R_{balance}$. To the best of the authors' knowledge, this is the first method to greatly reduce gender bias in diffusion models without additional data, full re-training or hard prompt modification. We also explored the possibility of stabilizing more the fine-tune process using an alternative trust region constraint but the outcome was not improved.
\\

\noindent Our findings highlight the potential of RLAIF in fine-tuning image generation models for bias reduction. Nevertheless, this work is just a starting point, and further research is needed to generalize these findings and explore the extension of our methodology to address other forms of bias and enhance the robustness of the RLAIF pipeline. Further works could extend this method to different bias such as racial or cultural ones. This work was also limited to one-prompt results due to the limitation of our hardware in generating large batches, but future works could explore further multi-prompts fine-tuning. Overall, our project contributes to the ongoing efforts in developing AI systems that prioritize fairness, inclusivity, and responsible deployment.

\section{Code Availability}
Our codes are available at Github repository(\url{https://github.com/X-Chen97/cs285-proj}).

\section*{Declaration of Contribution}
The authors declare no conflicts of interest. All the team members contribute to this project equally. Virgile proposed the reward function and Xin implemented the model. Both authors participated into debugging, revision, and report writing. 

\section*{Acknowledgement}
This work was conducted in the context of Sergey Levine's Deep Reinforcement Learning course at UC Berkeley. We express our gratitude to Professor Levine and the class staff for their guidance.
The authors also want to thank Bakar Haghighi for participating in the proposal and milestone work, as well as his efforts in querying the classification model.

\bibliographystyle{IEEEtran}
\bibliography{references.bib}





\end{document}